\documentclass{article}

\usepackage[preprint]{corl_2026}
\usepackage{amsfonts} 
\usepackage{amsmath}
\usepackage{booktabs}
\usepackage{graphicx}
\usepackage{multirow}
\usepackage{caption}
\usepackage{colortbl}
\usepackage{wrapfig}
\usepackage{xcolor}
\usepackage{algorithm}
\usepackage{algpseudocode}
\usepackage{longtable}
\usepackage{float}

\title{Light-WAM: Efficient World Action Models with State-Fusion Action Decoding}

\author{
\normalfont
\textbf{Ziang Li}\textsuperscript{1,2}\thanks{Equal contribution.},
\textbf{Dongzhou Cheng}\textsuperscript{2,3}\footnotemark[1],
\textbf{Yibin Wang}\textsuperscript{2,4},
\textbf{Shiyue Wang}\textsuperscript{2,5}, \\
\textbf{Xiaoyang Xu}\textsuperscript{1},
\textbf{Lingxuan Weng}\textsuperscript{5},
\textbf{Juan Wang}\textsuperscript{1}\thanks{Corresponding authors.},
\textbf{Jiaqi Wang}\textsuperscript{2}\footnotemark[2]
\\[6pt]
{\small \textsuperscript{1} Wuhan University \quad
\textsuperscript{2} Shanghai Innovation Institute}\\
{\small \textsuperscript{3} Southeast University \quad
\textsuperscript{4} Fudan University \quad
\textsuperscript{5} East China Normal University}
}

\begin{document}

\hypersetup{linkcolor=black}
\maketitle
\hypersetup{
    colorlinks=true,
    linkcolor=red,
    urlcolor=magenta
}

\vspace{-2.0em}

\begin{abstract}
World Action Models (WAMs) extend robot policy learning by incorporating future prediction as an additional training objective, encouraging the policy to encode task-relevant temporal structure in its representations. Current WAMs often rely on large-scale generative architectures that incur high training costs and inference latency, making them difficult to deploy as efficient closed-loop policies. 
We propose \textbf{Light-WAM}, a lightweight World Action Model for efficient robot manipulation. Specifically, it is built with a compact video backbone and performs future-video supervision in a downsampled latent space, reducing the cost of video co-training while retaining its benefits for representation learning. 
For action prediction, Light-WAM introduces the StateFusionActionExpert, which reads adapted states from multiple backbone layers, fuses them through learned-query pooling, and directly predicts action chunks in a single forward pass. 
This design provides an efficient interface between video backbone representations and robot actions, avoiding the need for heavy generative action experts. 
Experiments demonstrate that Light-WAM maintains strong performance on LIBERO and achieves usable multi-task performance on RoboTwin 2.0, while using only $0.44$B trainable parameters. 
It also achieves $72.03$ms inference latency with $4.1$GiB peak GPU memory and improved training throughput. The code is available at \url{https://github.com/L1ziang/Light-WAM}.
\end{abstract}



\section{Introduction}

Vision Language Action (VLA) models have shown strong performance in instruction-following robot manipulation by mapping visual observations and language instructions to robot actions~\cite{brohan2022rt,zitkovich2023rt,kim2024openvla,o2024open,black2024pi_0,shukor2025smolvla}. 
World Action Models (WAMs) extend this formulation by training robot policies jointly with future video prediction~\cite{liang2025video,li2025unified,zhu2025unified,bi2025motus,li2026causal,yuan2026fast}. 
The future-video objective provides additional supervision on how the scene changes over time, enabling the policy to learn representations that capture object motion, interaction dynamics, and task progress. 
However, current WAMs typically couple future-video prediction and action generation within large-scale generative architectures, resulting in substantial GPU memory usage, training cost, and inference latency. These overheads make it challenging to deploy WAMs as efficient closed-loop robot policies.

Recent work has shown that test-time future video generation is not necessary for strong policy performance, suggesting that the main benefit of video prediction may come from training-time representation learning~\cite{yuan2026fast}. 
Building on this, we study whether WAMs can be made more efficient while retaining the training benefit of future-video prediction.
This leads to a compact WAM designed for efficient training and fast inference.

We propose \textbf{Light-WAM}, a lightweight World Action Model for efficient robot manipulation. 
Light-WAM uses Wan2.1-T2V-1.3B as the video backbone~\cite{wan2025wan}, keeps the pretrained backbone frozen, and adapts it with lightweight modules. 
To reduce the cost of video supervision, Light-WAM applies the future-video objective in a downsampled latent space. 
During inference, Light-WAM predicts action chunks from the current observation, without test-time future-video generation or a generative action expert. To connect the video backbone to robot actions, we introduce the StateFusionActionExpert. 
This module reads adapted states from multiple backbone layers and compresses dense video tokens with learned-query pooling. 
The pooled states are fused and mapped to actions in a single forward pass. 
This provides an efficient interface between video representations and action prediction, while allowing the action decoder to use information from different levels of the video backbone.

We evaluate Light-WAM on LIBERO~\cite{liu2023libero} and RoboTwin 2.0~\cite{chen2025robotwin}. 
On LIBERO, Light-WAM achieves $97.2\%$ average success without embodied pretraining, which is competitive with larger WAM baselines. 
On RoboTwin 2.0, Light-WAM achieves $76.4\%$ average success across 50 tasks. 
Compared with Fast-WAM~\cite{yuan2026fast}, Light-WAM reduces trainable parameters from $6.02$B to $0.44$B, improves training throughput by $4.25\times$, and reduces inference latency to $72.03$ms with $4.1$GiB peak GPU memory. 
These results show that Light-WAM substantially improves the efficiency of the WAM pipeline while maintaining strong LIBERO performance and achieving usable multi-task performance in the more challenging RoboTwin 2.0.

Our contributions are summarized as follows:
\begin{itemize}
    \item We propose Light-WAM, a lightweight World Action Model that combines a compact video backbone with downsampled latent-space video supervision, reducing the cost of WAM training while retaining the representation benefits of future-video co-training.

    \item We introduce the StateFusionActionExpert, a direct action decoder that bridges video backbone representations and robot actions. It fuses multi-level adapted states through learned-query pooling and predicts action chunks in a single forward pass.
    
    \item We evaluate Light-WAM on LIBERO and RoboTwin 2.0. 
    Light-WAM achieves strong LIBERO performance and usable multi-task performance on RoboTwin 2.0, while substantially reducing both training and inference costs compared with heavier WAM baselines.
\end{itemize}

\section{Related Work}

\paragraph{Vision Language Action Models.}
Vision Language Action (VLA) models have become a central paradigm for instruction-following robot manipulation. Given visual observations and a language instruction, these models predict robot actions, enabling task conditioning and scalable learning from multi-task robot datasets~\cite{brohan2022rt,zitkovich2023rt,kim2024openvla,black2024pi_0,shukor2025smolvla,bjorck2025gr00t,team2025gemini,intelligence2025pi_,kim2025fine,liu2025rdt,wang2026vla}. Recent work further improves the practicality of VLA policies: SmolVLA~\cite{shukor2025smolvla} explores compact architectures for efficient training and deployment, while VLA-Adapter~\cite{wang2026vla} introduces a lightweight interface for adapting vision-language representations to action prediction. However, these methods are primarily trained through action supervision, leaving the temporal structure of the task to be captured implicitly by the policy.

\paragraph{World Action Models.}
World Action Models (WAMs) provide a different perspective by coupling robot action learning with future video prediction. The future-video objective offers a temporal training signal that encourages the backbone to encode object motion, interaction dynamics, and task progress, leading to more world-aware visual representations~\cite{liang2025video,li2025unified,zhu2025unified,bi2025motus,li2026causal,yuan2026fast,du2023learning,wu2024unleashing,bharadhwaj2024gen2act,zhou2024robodreamer,hu2024video,liao2025genie,ye2026world}. Recent WAM systems such as Motus~\cite{bi2025motus}, LingBot-VA~\cite{li2026causal}, and Fast-WAM~\cite{yuan2026fast} demonstrate the value of video co-training for robot policy learning, but they often rely on large video-action generative architectures and expensive training or inference pipelines.

Our work shares the efficiency-oriented goal of recent VLA policies, but targets WAM robot policies, where future video prediction is used to shape the visual representations for robot control. It is also closely related to Fast-WAM, which shows that the video prediction branch can be used as training-time supervision without being executed during inference. 
Rather than focusing on inference-time video rollout, we focus on improving the efficiency of the overall WAM pipeline.

\section{Methodology}

\subsection{Overview}

Building on this motivation, Light-WAM keeps future-video supervision during training, but instantiates the policy with a compact backbone and a direct action interface at test time. Given the current observation $o$, language instruction $l$, and proprioceptive state $p$, it predicts an action sequence by
\begin{equation}
    \hat{A}
    =
    \pi_{\phi}
    \left(
    h_{\theta}(o,l,p)
    \right),
\end{equation}
where $h_{\theta}$ denotes the multi-level backbone representation extracted from the adapted video backbone, and $\pi_{\phi}$ is the StateFusionActionExpert. 
Light-WAM is designed to make the WAM pipeline efficient in both training and inference. 
It uses a compact video backbone with minimal adaptation to preserve the pretrained video prior, reduces video supervision cost via latent-space downsampling, and employs the StateFusionActionExpert to decode actions directly from multi-level backbone states, enabling fast closed-loop execution without iterative action denoising. 
The overall architecture of Light-WAM is illustrated in Figure~\ref{fig:method_overview}.
Detailed procedures are provided in Appendix~\ref{app:algorithmic_details}.

\begin{figure*}[t]
    \centering
    \includegraphics[width=0.98\textwidth]{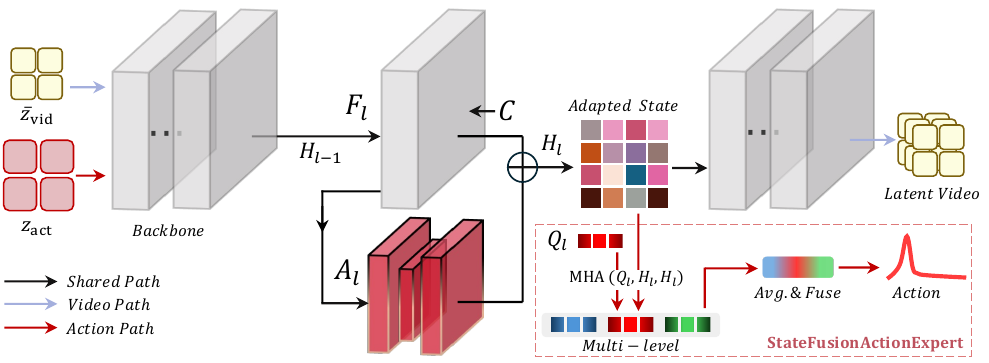}
    \caption{
    \textbf{Overview of Light-WAM. }
    Light-WAM shares an adapted video backbone between video co-training and action prediction. 
    During training, the \textcolor[RGB]{124,134,178}{video branch} applies future-video supervision to downsampled latent videos $\bar{z}_{\mathrm{vid}}$, reducing the token cost of temporal supervision. 
    The \textcolor[RGB]{191,13,16}{action prediction branch} runs in both training and inference: it takes the current observation latent $z_{\mathrm{act}}$ and predicts action chunks without future-video rollout. 
    The backbone is adapted with LoRA and sparse WAM adapters, and multi-level adapted states are fused by the StateFusionActionExpert through learned-query pooling for single-pass action decoding.
    }
    \label{fig:method_overview}
\end{figure*}

\subsection{Video Backbone Adaptation}

Light-WAM uses Wan2.1-T2V-1.3B as the video backbone~\cite{wan2025wan}. 
Given a VAE latent input $z$, the patch embedding layer produces the initial video-token state:
\begin{equation}
    H_0 = \mathrm{PatchEmbed}(z) \in \mathbb{R}^{B \times N \times d},
\end{equation}
where $N$ is the number of spatiotemporal video tokens and $d$ is the hidden dimension. 
The language instruction is encoded as text context tokens, and the proprioceptive state is projected to the same context dimension and appended to them:
\begin{equation}
    C = [c_1,\ldots,c_L,c_{\mathrm{prop}}].
\end{equation}
The backbone then updates the video-token state through transformer blocks, with $C$ provided as the cross-attention context. 
To preserve the pretrained video prior, we freeze the original Wan backbone and adapt it through low-rank updates on its attention and feed-forward projections~\cite{hu2022lora}. 
We further insert lightweight WAM adapters at a sparse set of backbone depths. Let $F_{\ell}$ denote the $\ell$-th transformer block and $A_{\ell}$ the WAM adapter inserted at that depth, if present. 
Given the previous hidden state $H_{\ell-1}$ and context $C$, the layer update is
\begin{equation}
    U_{\ell}=F_{\ell}(H_{\ell-1},C),
    \qquad
    H_{\ell}=
    \begin{cases}
        U_{\ell}+A_{\ell}(U_{\ell}), & \ell \in \mathcal{I},\\
        U_{\ell}, & \text{otherwise},
    \end{cases}
\end{equation}
where $\mathcal{I}$ denotes the depths exposed to the action decoder and $A_{\ell}$ is a lightweight bottleneck MLP. In this way, low-rank updates provide lightweight adaptation across the backbone, while the sparse WAM adapters provide additional robot-domain adaptation capacity at selected depths. The action branch reads a sparse set of adapted backbone states:
\begin{equation}
    \mathcal{H} = \{H_{\ell}\}_{\ell \in \mathcal{I}}.
\end{equation}
These multi-level backbone states form the interface between the video backbone and the StateFusionActionExpert. 
Instead of using only the final representation or exposing all backbone activations, Light-WAM selects a small set of states from different backbone levels, allowing the action head to access visual information at multiple granularities while keeping action decoding efficient.

\subsection{Efficient Latent Video Co-training}

The future-video branch provides temporal supervision during training. 
Let $G_{\theta}^{\mathrm{vid}}$ denote the video prediction branch, which includes the adapted video backbone and the final video prediction head. 
Let $\bar{z}_{\mathrm{vid}}=D(z_{\mathrm{vid}})$ be the latent video after spatial downsampling, and let $\bar{z}_{t}$ denote its flow-matching perturbation at time $t$. 
The video branch is optimized by
\begin{equation}
    \mathcal{L}_{\mathrm{video}}
    =
    \left\|
    G_{\theta}^{\mathrm{vid}}(\bar{z}_{t}, t, C) - u_t
    \right\|_2^2,
\end{equation}
where $u_t$ is the corresponding flow-matching target~\cite{lipman2022flow}.
The first latent frame is also downsampled by $D(\cdot)$ and kept fixed in $\bar{z}_{t}$ as the observation condition. For action prediction, Light-WAM takes the current observation from the original-resolution latent video:
\begin{equation}
    z_{\mathrm{act}} = z_{\mathrm{vid}}^{(0)},
\end{equation}
and does not apply the additional spatial downsampling used for video supervision. 
Thus, the video branch learns future dynamics in a lower-cost downsampled latent space, while the action branch preserves the original-resolution current observation needed for manipulation.

\subsection{Query-Bottlenecked State Fusion and Action Decoding}

Given the full-resolution current observation latent $z_{\mathrm{act}}$, Light-WAM runs the adapted video backbone once and obtains the multi-level backbone states $\mathcal{H}=\{H_{\ell}\}_{\ell \in \mathcal{I}}$. 
The StateFusionActionExpert converts these dense video-token states into a fixed-width action state through learned-query pooling. 
This design is related to prior query-based pooling methods that use learnable queries to compress dense input tokens into compact representations~\cite{lee2019set,li2023blip}. 
For each backbone state $H_{\ell} \in \mathcal{H}$, we learn a set of query tokens
\begin{equation}
    Q_{\ell} \in \mathbb{R}^{N_q \times d}.
\end{equation}

The query tokens attend to the video tokens of the corresponding backbone level, producing $P_{\ell}\in\mathbb{R}^{B\times N_q\times d}$, which is then averaged over queries and normalized:
\begin{equation}
    P_{\ell}=\mathrm{MHA}(Q_{\ell},H_{\ell},H_{\ell}), 
    \qquad
    s_{\ell}=\mathrm{LN}\left(\frac{1}{N_q}\sum_{j=1}^{N_q}P_{\ell,j}\right).
\end{equation}

$\mathrm{MHA}$ and $\mathrm{LN}$ denote multi-head attention~\cite{vaswani2017attention} and layer normalization~\cite{ba2016layer}, respectively.
The number of queries controls the information passed from the video backbone to the action head. 
Too few queries may lose manipulation-relevant visual details, while too many reduce the compression effect and increase the burden on the action decoder. 
This design provides a controlled bottleneck that compresses dense video tokens into level-wise state representations. The resulting states are projected and fused as
\begin{equation}
    h =
    \phi_{\mathrm{trunk}}
    \left(
    \phi_{\mathrm{fuse}}
    \left(
    [M_{\ell}(s_{\ell})]_{\ell \in \mathcal{I}}
    \right)
    \right).
\end{equation}
To decode actions, Light-WAM uses step embeddings $\{e_k\}_{k=1}^{K}$, where $K$ is the action horizon. Each embedding is projected by $\psi(\cdot)$ and added to the fused state $h$, after which an output head predicts the corresponding action:
\begin{equation}
    r_k = h + \psi(e_k), 
    \qquad
    \hat{a}_k = \phi_{\mathrm{out}}\left(\mathrm{LN}(r_k)\right),
    \qquad
    \hat{A}=[\hat{a}_1,\ldots,\hat{a}_K]\in\mathbb{R}^{B\times K\times d_a}.
\end{equation}

The full training objective combines future-video supervision and action regression:
\begin{equation}
    \mathcal{L}
    =
    \mathcal{L}_{\mathrm{video}}
    +
    \lambda 
    \mathcal{L}_{\mathrm{action}}(\hat{A}, A),
\end{equation}
where $A$ denotes the target action sequence and $\mathcal{L}_{\mathrm{action}}$ measures the regression error. 
At inference time, Light-WAM directly predicts actions from the current observation without future-video rollout.

\section{Experiments}

\subsection{Experimental Setup}

\paragraph{Benchmarks and data.}
We evaluate Light-WAM on LIBERO~\cite{liu2023libero} and RoboTwin 2.0~\cite{chen2025robotwin}. 
For LIBERO, we use the official datasets and report success rates on four suites: Spatial, Object, Goal, and Long. 
For RoboTwin 2.0, we follow the multi-task evaluation protocol used in prior work~\cite{bi2025motus,li2026causal,yuan2026fast}: one policy is trained on 50 tasks using 2,500 clean demonstrations and 25,000 randomized demonstrations.
We report performance under both clean and randomized evaluation settings.

\paragraph{Implementation details.}
Light-WAM uses Wan2.1-T2V-1.3B~\cite{wan2025wan} as a frozen video backbone and trains only the lightweight adaptation and action prediction modules. 
We insert WAM adapters at layers $\{8,16,24\}$, set the number of learned queries to $16$ for each selected layer, and apply $2\times$ spatial latent downsampling in the video co-training branch. 
The default model has $1.99$B total parameters and $0.44$B trainable parameters. 
We train with AdamW using learning rate $1e-4$, weight decay $1e-2$, LIBERO batch size $64$, and RoboTwin 2.0 batch size $128$. 
Training is conducted on $4$ NVIDIA H100 GPUs, and inference is measured on NVIDIA RTX 4090 48G GPUs. 
Additional implementation details are provided in Appendix~\ref{app:training_details}.

\paragraph{Baselines and metrics.}
We compare Light-WAM with representative VLA and WAM policies, including OpenVLA~\cite{kim2024openvla}, OpenVLA-OFT~\cite{kim2025fine}, VLA-Adapter~\cite{wang2026vla}, $\pi_0$~\cite{black2024pi_0}, $\pi_{0.5}$~\cite{intelligence2025pi_}, X-VLA~\cite{zheng2025x}, Motus~\cite{bi2025motus}, LingBot-VA~\cite{li2026causal}, and Fast-WAM~\cite{yuan2026fast}. 
Since large-scale embodied pretraining can significantly affect downstream manipulation performance, we indicate whether each method uses it. 
For task performance, we report success rate. 
For efficiency, we report trainable parameters, training throughput, inference latency, and peak GPU memory.

\subsection{LIBERO Results}

Table~\ref{tab:libero} reports LIBERO results. 
Light-WAM achieves $97.2\%$ average success, ranking first among methods without embodied pretraining and third among all compared methods. 
This indicates that Light-WAM remains competitive on LIBERO with fewer parameters than existing WAM baselines.

Light-WAM obtains $98.2\%$, $99.6\%$, $97.8\%$, and $93.0\%$ on Spatial, Object, Goal, and Long, respectively. 
The Long suite remains the most challenging setting, where larger policies such as Motus and LingBot-VA achieve higher success rates, suggesting that long-horizon tasks can still benefit from larger model capacity. 
Overall, the LIBERO results show that Light-WAM achieves competitive task performance with improved model efficiency.

\begin{table*}[t]
\centering
\caption{
\textbf{LIBERO success rates on the four official suites.} 
We report average success, rank among methods without embodied pretraining (w/o EPT), and overall rank.
}
\label{tab:libero}
\setlength{\tabcolsep}{3.5pt}
\resizebox{\textwidth}{!}{
\begin{tabular}{llccccccccc}
\toprule
\multirow{2}{*}{Type} 
& \multirow{2}{*}{Method} 
& \multirow{2}{*}{Params} 
& \multirow{2}{*}{EPT} 
& \multirow{2}{*}{Spatial} 
& \multirow{2}{*}{Object} 
& \multirow{2}{*}{Goal} 
& \multirow{2}{*}{Long} 
& \multirow{2}{*}{Avg.} 
& w/o EPT 
& Overall \\
& & & & & & & & & Rank & Rank \\
\midrule
\multirow{5}{*}{VLA} 
& OpenVLA~\cite{kim2024openvla} & 7B & w/ & 84.7 & 88.4 & 79.2 & 53.7 & 76.5 & -- & 9 \\
& OpenVLA-OFT~\cite{kim2025fine} & 7B & w/ & 97.6 & 98.4 & 97.9 & 94.5 & 97.1 & -- & 4 \\
& VLA-Adapter~\cite{wang2026vla} & 0.6B & w/o & 96.0 & 96.8 & 97.4 & 94.4 & 96.2 & 3 & 7 \\
& $\pi_0$~\cite{black2024pi_0} & 3B & w/ & 96.8 & 98.8 & 95.8 & 85.2 & 94.1 & -- & 8 \\
& $\pi_{0.5}$~\cite{intelligence2025pi_} & 3B & w/ & 98.8 & 98.2 & 98.0 & 92.4 & 96.9 & -- & 6 \\
\midrule
\multirow{4}{*}{WAM} 
& Motus~\cite{bi2025motus} & 8B & w/ & 96.8 & 99.8 & 96.6 & 97.6 & 97.7 & -- & 2 \\
& LingBot-VA~\cite{li2026causal} & 5.3B & w/ & 98.5 & 99.6 & 97.2 & 98.5 & 98.5 & -- & \textbf{1} \\
& Fast-WAM~\cite{yuan2026fast} & 6B & w/o & 97.0 & 99.4 & 96.6 & 94.8 & 97.0 & 2 & 5 \\
\rowcolor[HTML]{F2F2F2} & \textbf{Light-WAM} & 2B & w/o & 98.2 & 99.6 & 97.8 & 93.0 & 97.2 & \textbf{1} & 3 \\
\bottomrule
\end{tabular}
}
\end{table*}

\subsection{Multi-Task Learning on RoboTwin 2.0}

We further evaluate Light-WAM on RoboTwin 2.0 to study whether the lightweight architecture remains usable in a larger multi-task setting. 
Unlike LIBERO, RoboTwin 2.0 requires a single policy to learn across 50 bimanual manipulation tasks and handle randomized visual and physical conditions. 
This setting is more challenging for a lightweight model such as Light-WAM, with only $0.44$B trainable parameters and a direct action head rather than large generative action experts.

\begin{figure*}[t]
\centering

\begin{minipage}[t]{0.58\textwidth}
\vspace{0pt}
\captionof{table}{
\textbf{RoboTwin 2.0 success rates on 50 tasks. }
We report clean, randomized, and average success.
}
\label{tab:robotwin}
\setlength{\tabcolsep}{3.5pt}
\resizebox{\linewidth}{!}{
\begin{tabular}{llccccc}
\toprule
Type & Method & Params & EPT & Clean & Randomized & Avg. \\
\midrule
\multirow{3}{*}{VLA}
& $\pi_0$~\cite{black2024pi_0} & 3B & w/ & 65.9 & 58.4 & 62.2 \\
& $\pi_{0.5}$~\cite{intelligence2025pi_} & 3B & w/ & 82.7 & 76.8 & 79.8 \\
& X-VLA~\cite{zheng2025x} & 0.9B & w/ & 72.9 & 72.8 & 72.9 \\
\midrule
\multirow{6}{*}{WAM}
& Motus~\cite{bi2025motus} & 8B & w/ & 88.7 & 87.0 & 87.8 \\
& Motus~\cite{bi2025motus} & 8B & w/o & 72.8 & 77.0 & 74.9 \\
& LingBot-VA~\cite{li2026causal} & 5.3B & w/ & 92.9 & 91.5 & 92.2 \\
& LingBot-VA~\cite{li2026causal} & 5.3B & w/o & 80.6 & -- & 80.6 \\
& Fast-WAM~\cite{yuan2026fast} & 6B & w/o & 91.9 & 91.8 & 91.9 \\
\rowcolor[HTML]{F2F2F2} & \textbf{Light-WAM} & 2B & w/o & 76.4 & 76.3 & 76.4 \\
\bottomrule
\end{tabular}
}
\end{minipage}
\hfill
\begin{minipage}[t]{0.38\textwidth}
\vspace{5pt}
\centering
\makebox[\linewidth][r]{%
  \includegraphics[width=1.11\linewidth]{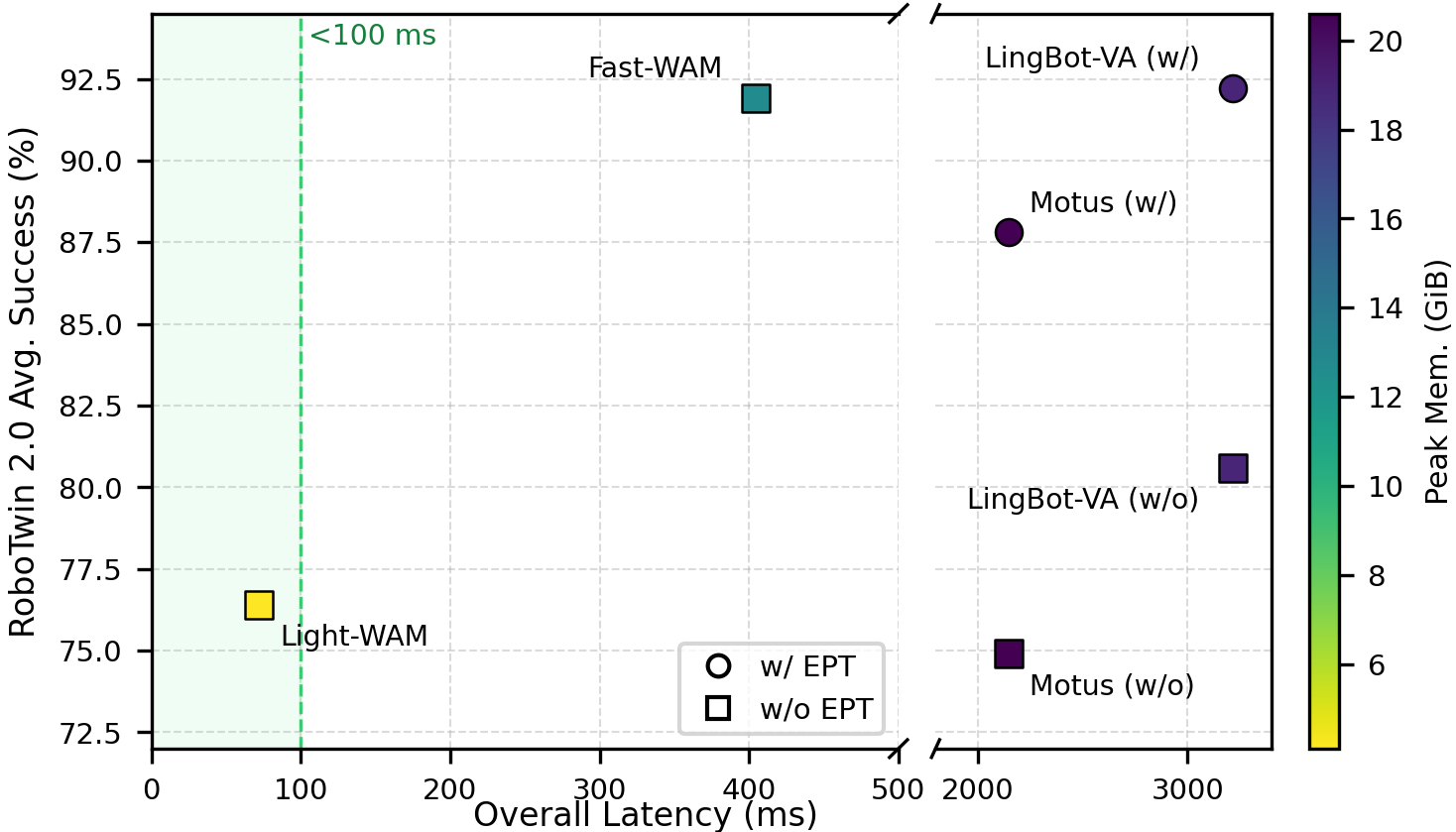}%
}
\hspace*{-0.08\textwidth}%
\begin{minipage}{1.05\linewidth}
\vspace{4pt}
\captionof{figure}{
RoboTwin 2.0 inference efficiency-performance comparison.
}
\label{fig:robotwin_results}
\end{minipage}
\end{minipage}

\end{figure*}

As shown in Table~\ref{tab:robotwin}, Light-WAM achieves $76.4\%$ average success on RoboTwin 2.0 without embodied pretraining. 
Although it does not match Fast-WAM or the strongest embodied-pretrained WAMs, this result shows that Light-WAM can obtain usable multi-task performance with a much smaller trainable parameter budget. 
It outperforms $\pi_0$ and X-VLA in this comparison, and is competitive with Motus without embodied pretraining. 
These results position Light-WAM as an efficient WAM policy. While larger models perform better in the more complex RoboTwin 2.0 setting, Light-WAM achieves usable multi-task performance with much lower training and inference cost. Figure~\ref{fig:robotwin_results} visualizes the inference-side trade-off: Light-WAM achieves much lower inference latency and peak GPU memory among WAM methods, while maintaining usable average success on RoboTwin 2.0.

\subsection{Efficiency Analysis}

Light-WAM is designed to reduce the cost of the entire WAM training and inference pipeline. 
Table~\ref{tab:training_efficiency} reports training efficiency. 
Compared with Fast-WAM, Light-WAM reduces total training-time parameters from $6.73$B to $1.99$B and trainable parameters from $6.02$B to $0.44$B, corresponding to $3.4\times$ and $13.7\times$ reductions, respectively. 
Peak per-GPU memory decreases from $70.7$GiB to $43.1$GiB, and throughput increases from $0.49$ to $2.08$ steps/s.

We further analyze the contribution of each efficiency component. 
A compact video backbone alone does not guarantee faster training, partly because the Wan2.1 VAE produces a denser latent grid than the high-compression VAE used by Wan2.2-TI2V-5B~\cite{wan2025wan}.
Introducing the StateFusionActionExpert reduces the action-side parameter and computation cost. 
Latent caching removes online VAE encoding from the training loop, and $2\times$ spatial downsampling reduces the token cost of future-video co-training. 
Together, these choices reduce training cost while preserving future-video supervision as part of the learning objective.

\begin{table*}[t]
\centering
\caption{
\textbf{Training efficiency analysis.} We measure on $4\times$ NVIDIA H100 GPUs with effective global batch size 64. 
Loaded Params denotes training-time loaded model parameters. 
}
\label{tab:training_efficiency}
\setlength{\tabcolsep}{3.5pt}
\resizebox{\textwidth}{!}{
\begin{tabular}{@{}lccccccccc@{}}
\toprule
Model / Variant 
& Action Head 
& \shortstack{Latent\\Cache} 
& \shortstack{Video\\Downsample} 
& \shortstack{Loaded\\Params} 
& \shortstack{Trainable\\Params} 
& \shortstack{Mem.\\ / GPU} 
& Samples/s 
& Steps/s 
& \shortstack{Steps/s\\Norm.} \\
\midrule
Fast-WAM~\cite{yuan2026fast} 
& DiT~\cite{peebles2023scalable} & No & $1\times$ & 6.73B & 6.02B & 70.7GiB & 31.6 & 0.49 & $1.00\times$ \\

Light-WAM 
& DiT~\cite{peebles2023scalable} & No & $1\times$ & 2.28B & 0.73B & 58.9GiB & 27.7 & 0.43 & $0.88\times$ \\

Light-WAM$^\ast$
& StateFusion & No & $1\times$ & 1.99B & 0.44B & 48.6GiB & 35.7 & 0.56 & $1.14\times$ \\

Light-WAM$^\ast$
& StateFusion & Yes & $1\times$ & 1.99B & 0.44B & 48.2GiB & 55.3 & 0.86 & $1.76\times$ \\

\textbf{Light-WAM} 
& StateFusion & Yes & $2\times$ & 1.99B & 0.44B & 43.1GiB & 133.3 & 2.08 & $\mathbf{4.25\times}$ \\
\bottomrule
\end{tabular}
}
\vspace{0.2em}
\hspace*{-2.75cm}{
\footnotesize{
$^\ast$ These variants use batch size 8 per GPU and gradient accumulation 2 to avoid OOM.
}
}
\end{table*}

Table~\ref{tab:inference_efficiency} reports inference efficiency on RoboTwin 2.0 inputs. 
Latency is measured per action query with cached language context, including VAE encoding and policy forward. 
Light-WAM achieves $72.03$ms overall latency with $4.1$GiB peak GPU memory, substantially lower than prior WAM methods. 
The breakdown shows that its action branch takes only $2.1$ms, while larger WAMs spend much more time on iterative action prediction or joint video-action generation. 
These results show that Light-WAM enables fast and memory-efficient action prediction for closed-loop control.

\begin{table*}[t]
\centering
\caption{
\textbf{Inference efficiency on RoboTwin 2.0 inputs. }
Latency is measured per action query with cached language context on a single NVIDIA RTX 4090 48GB GPU. 
Overall latency includes VAE encoding and policy forward, while simulator and I/O overheads are excluded.
}
\label{tab:inference_efficiency}
\setlength{\tabcolsep}{3.5pt}
\resizebox{\textwidth}{!}{
\begin{tabular}{lccccccccc}
\toprule
Model & Params & Prediction Scope & VAE Enc. & Visual Branch & Action Branch & Policy Forward & Peak Mem. & Overall Latency & Norm. \\
\midrule
$\pi_{0.5}$~\cite{intelligence2025pi_} 
& 3B & Action-only & -- & -- & -- & -- & $>8$GiB & 76ms$^\ast$ & $1.00\times$ \\

LingBot-VA~\cite{li2026causal} 
& 5.3B & Video + action & 223.8ms & -- & -- & 2990.1ms & 18.9GiB & 3214.14ms & $42.29\times$ \\

Motus~\cite{bi2025motus} 
& 8B & Video + action & 16.1ms & -- & -- & 2130.7ms & 20.6GiB & 2148.68ms & $28.27\times$ \\

Fast-WAM~\cite{yuan2026fast} 
& 6B & Action-only & 11.3ms & 36.0ms & 356.8ms & 392.8ms & 12.7GiB & 404.62ms & $5.32\times$ \\

\rowcolor[HTML]{F2F2F2} \textbf{Light-WAM} 
& 2B & Action-only & 12.7ms & 56.5ms & 2.1ms & 58.6ms & \textbf{4.1GiB} & \textbf{72.03ms} & $\mathbf{0.95\times}$ \\
\bottomrule
\end{tabular}
}
\vspace{0.2em}
\hspace*{-1.9cm}{
\footnotesize{
$^\ast$ The $\pi_{0.5}$ latency is reported by~\cite{black2026real}, which also evaluates on an NVIDIA RTX 4090 GPU.
}
}
\end{table*}

\subsection{Ablation Studies}

We conduct ablations on LIBERO-Spatial for three Light-WAM designs: the resolution of video co-training, the number of adapter layers, and the capacity of learned-query pooling. 
As shown in Table~\ref{tab:ablation}, using the original-resolution video latent for co-training improves success from $98.2\%$ to $99.0\%$. 
This suggests that higher-resolution video supervision can further improve policy performance. 
However, full-resolution video co-training raises the training cost substantially, as shown in Table~\ref{tab:training_efficiency}. 
We therefore use $2\times$ latent downsampling to balance performance and training efficiency.

Increasing the number of adapter layers from $3$ to $5$ gives similar performance, with success changing from $98.2\%$ to $98.0\%$. 
This indicates that adding more adapter layers brings no clear gain in this setting. 
Considering the additional parameters and computation, we choose a sparse three-layer configuration $\{8,16,24\}$, which provides multi-level representations for the action decoder. 
Finally, reducing the number of learned queries from $16$ to $8$ decreases success to $95.4\%$, suggesting that the query bottleneck needs sufficient capacity to preserve manipulation-relevant visual information.

\begin{wraptable}{l}{0.55\textwidth}
\vspace{-1.1em}
\centering
\caption{
\textbf{Ablations on LIBERO-Spatial. }
The default Light-WAM uses $2\times$ latent downsampling, adapters at layers $\{8,16,24\}$, and 16 learned queries.
}
\label{tab:ablation}
\setlength{\tabcolsep}{3pt}
\scriptsize
\renewcommand{\arraystretch}{1.35}
\resizebox{\linewidth}{!}{
\begin{tabular}{lcccc}
\toprule
Variant & DS & Adapter Layers & Queries & Success \\
\midrule
\textbf{Light-WAM} & $2\times$ & $\{8,16,24\}$ & 16 & 98.2 \\
w/o downsample & $1\times$ & $\{8,16,24\}$ & 16 & 99.0 \\
w/ 5 adapter layers & $2\times$ & $\{4,8,16,20,24\}$ & 16 & 98.0 \\
w/ 8 learned queries & $2\times$ & $\{8,16,24\}$ & 8 & 95.4 \\
\bottomrule
\end{tabular}
}
\vspace{-1.0em}
\end{wraptable}

\textbf{Overall}, Light-WAM achieves competitive LIBERO performance, usable 50-task performance on RoboTwin 2.0, and much lower training and inference cost. 
These results support the main design of Light-WAM: future-video prediction is retained as downsampled latent-space supervision, while multi-level adapted backbone states are fused by a single-pass StateFusionActionExpert for action decoding.

\subsection{Qualitative Analysis}

\begin{figure*}[t]
    \centering
    \includegraphics[width=0.99\textwidth]{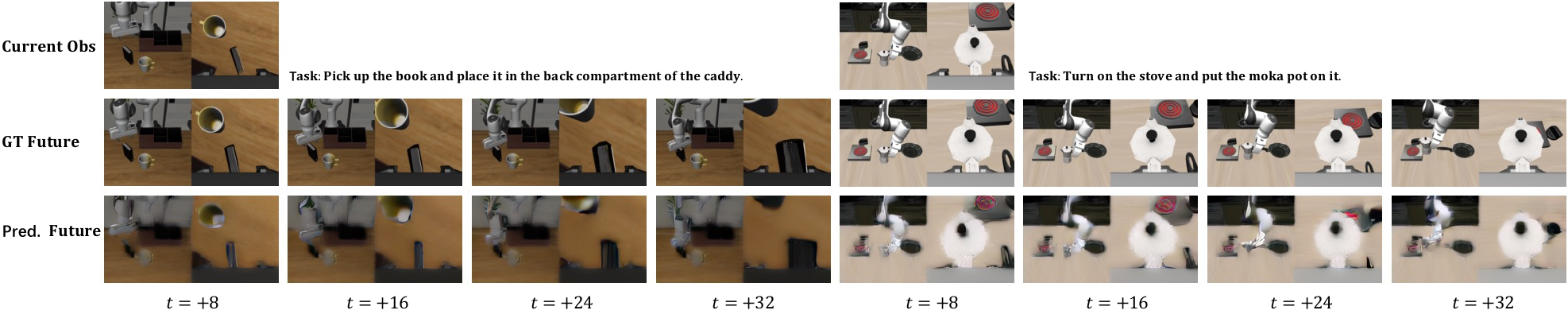}
    \vspace{-0.1em}

    \includegraphics[width=0.99\textwidth]{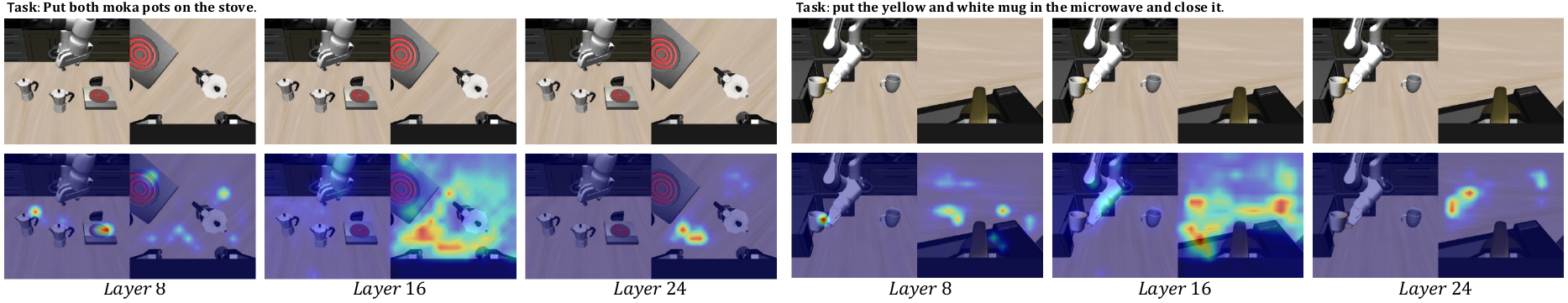}
    \vspace{-0.5em}
    \caption{
    \textbf{Qualitative analysis.}
    \small{Top: future-video predictions compared with reference rollout frames at $t=\{+8,+16,+24,+32\}$.
    Bottom: learned-query visualizations from the StateFusionActionExpert.}
    }
    \label{fig:qualitative}
    \vspace{-1.8em}
\end{figure*}

\vspace{-0.8em}
\paragraph{Future video visualization.}
The top row of Figure~\ref{fig:qualitative} shows examples from the video branch. 
For each task, we compare the predicted future frames with reference future frames from the environment rollout. 
The predictions are smoother than the reference frames because the video branch is trained in a downsampled latent space. 
However, they still capture the main motion and scene changes, suggesting that the video branch learns useful temporal information during training.

\vspace{-1.2em}
\paragraph{Learned-query visualization.}
The bottom row of Figure~\ref{fig:qualitative} visualizes attention maps derived from learned-query pooling. 
When projected back to image space, the maps from layers 8, 16, and 24 tend to emphasize different task-relevant regions, such as manipulated objects, the gripper, and target areas. 
This suggests that the selected backbone layers provide complementary visual cues, which is consistent with our design of fusing multi-level adapted states for action decoding.


\vspace{-0.8em}
\subsection{Real-World Evaluation}
\vspace{-0.8em}

\begin{figure}[H]
    \centering
    \includegraphics[width=\textwidth]{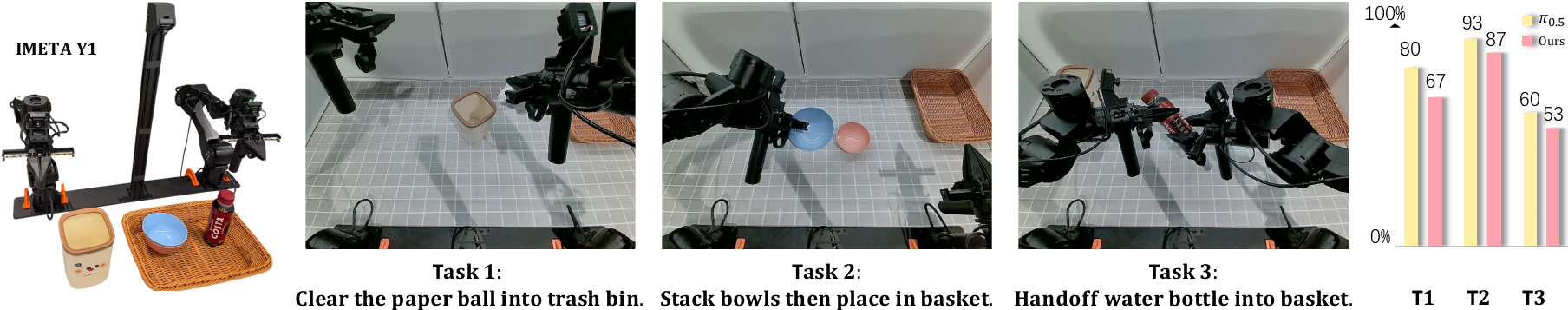}
    \caption{
    \textbf{Real-world evaluation.}
    Robot setup and success rates on three dual-arm tasks.
    }
    \label{fig:real_world}
\end{figure}

We evaluate Light-WAM on the IMETA Y1 dual-arm robot platform with three real-world manipulation tasks. 
For each task, we collect 50 demonstrations for training and compare Light-WAM with $\pi_{0.5}$ under the same setting. 
Figure~\ref{fig:real_world} shows the robot setup, task observations, and success rates for the three tasks. 
Additional rollout frames are provided in Appendix~\ref{app:real_world_rollouts}.

\vspace{-2.0em}
\section{Conclusion and Limitations}
\vspace{-0.4em}

We presented Light-WAM, a lightweight World Action Model for efficient robot manipulation. 
By combining a compact video backbone, downsampled latent-space video supervision, and the StateFusionActionExpert, Light-WAM improves the efficiency of both WAM training and inference. 
Experiments on LIBERO, RoboTwin 2.0, and real-world dual-arm tasks show a favorable performance-efficiency trade-off.
There are also several limitations.
In more challenging multi-task settings, larger WAMs and embodied-pretrained policies continue to achieve higher success rates, suggesting that model capacity and large-scale embodied data remain important for complex manipulation. 
Moreover, although we evaluate on existing benchmarks and real-world tasks, we do not train or test on benchmarks specifically designed for policy generalization and robustness, such as LIBERO-Plus~\cite{fei2025libero}. 
Future work will incorporate data augmentation and robustness-oriented training to further improve the generalization ability of Light-WAM.

\clearpage


\bibliography{example}  

\clearpage

\appendix

\section{Algorithmic Details}
\label{app:algorithmic_details}

\paragraph{Backbone adaptation.}
Light-WAM uses Wan2.1-T2V-1.3B as the video backbone and keeps the pretrained backbone weights frozen. 
We adapt the backbone with two lightweight components. 
First, LoRA is applied to the self-attention, cross-attention, and feed-forward projections of all backbone blocks. 
Second, sparse WAM adapters are inserted at layers $\{8,16,24\}$. 
Each WAM adapter is a residual bottleneck module:
\[
    A_{\ell}(x)
    =
    \gamma\,
    W_{\ell}^{\mathrm{up}}
    \sigma
    \left(
    W_{\ell}^{\mathrm{down}} x
    \right),
\]
where $W_{\ell}^{\mathrm{down}}$ maps the backbone hidden state to a 256-dimensional bottleneck, $W_{\ell}^{\mathrm{up}}$ maps it back to the backbone hidden dimension, and $\gamma$ is the adapter scale. 
For a selected layer $\ell$, the adapted state is computed as
\[
    H_{\ell}
    =
    U_{\ell}
    +
    A_{\ell}(U_{\ell}),
\]
where $U_{\ell}$ denotes the output of the corresponding backbone block. 
In our default configuration, $\gamma=1.0$. 
The adapted states from the selected layers are exposed to the StateFusionActionExpert for action prediction, while the final backbone output is used by the video prediction head for future-video co-training.

\paragraph{State-fusion.}
The StateFusionActionExpert maps the selected adapted backbone states to action chunks through query-based pooling and lightweight state fusion. 
For each selected layer $\ell\in\mathcal{I}$, we use a layer-specific set of learnable queries $Q_{\ell}$ to attend to the adapted video tokens $H_{\ell}$:
\[
    P_{\ell}
    =
    \mathrm{MHA}(Q_{\ell},H_{\ell},H_{\ell}),
    \qquad
    s_{\ell}
    =
    \mathrm{LN}
    \left(
    \frac{1}{N_q}
    \sum_{j=1}^{N_q}
    P_{\ell,j}
    \right).
\]
In our default configuration, each layer uses $N_q=16$ queries and 8 attention heads. 
The pooled state $s_{\ell}$ is projected to a 4608-dimensional feature, and the features from layers $\{8,16,24\}$ are concatenated and projected to a 6144-dimensional fused state. 
A single residual MLP block further processes the fused state. 
For temporal decoding, sinusoidal step-position embeddings of width 256 are projected and added to the fused state, after which an output MLP predicts the action at each step. 
For RoboTwin 2.0, the decoder outputs a $24\times14$ action chunk.

\begin{algorithm}[t]
\caption{Light-WAM training on RoboTwin 2.0}
\label{alg:lightwam_training}
\begin{algorithmic}[1]
\Require Observation sequence $I_{0:32}$, target action chunk $A=\{a_k\}_{k=0}^{K-1}$, language embedding $c$, proprioceptive states $p_{0:K-1}$
\Require Selected adapter layers $\mathcal{I}=\{8,16,24\}$
\State Construct the RoboTwin canvas video $V$ from the three camera streams, and subsample frames with stride 4:
\[
    V_{\mathrm{sub}}=[I_0,I_4,I_8,\ldots,I_{32}].
\]
\State Encode $V_{\mathrm{sub}}$ into Wan2.1 VAE latents $z_{\mathrm{vid}}$ using cached latents when available.
\State Build the cross-attention context
\[
    C=[c_1,\ldots,c_{128},c_{\mathrm{prop}}].
\]

\Statex \Comment{Future-video co-training branch}
\State Spatially downsample the video latents, $\bar{z}_{\mathrm{vid}}=D(z_{\mathrm{vid}})$, and sample a flow-matching timestep $t$ and noise $\epsilon$.
\State Construct the perturbed latent $\bar{z}_{t}$ from $\bar{z}_{\mathrm{vid}}$, while keeping the first latent frame fixed as the observation anchor.
\State Predict the flow target with the adapted video backbone and video head:
\[
    \hat{u}_{t}=G_{\theta}^{\mathrm{vid}}(\bar{z}_{t},t,C),
    \qquad
    \mathcal{L}_{\mathrm{video}}
    =
    \left\|
    \hat{u}_{t}-u_{t}
    \right\|_2^2 .
\]

\Statex \Comment{Action prediction branch}
\State Take the current observation latent at the original latent resolution:
\[
    z_{\mathrm{act}}=z_{\mathrm{vid}}^{(0)}.
\]
\State Run the adapted video backbone on $z_{\mathrm{act}}$ and collect multi-level adapted states:
\[
    \mathcal{H}
    =
    \{H_{\ell}\}_{\ell\in\mathcal{I}}
    =
    h_{\theta}(z_{\mathrm{act}},C).
\]
\State Predict the action chunk with the StateFusionActionExpert:
\[
    \hat{A}
    =
    \{\hat{a}_k\}_{k=0}^{K-1}
    =
    \pi_{\phi}(\mathcal{H}).
\]
\State Compute the weighted action regression loss:
\[
    \mathcal{L}_{\mathrm{action}}
    =
    \sum_{k=0}^{K-1}
    w_k
    \left\|
    \hat{a}_k-a_k
    \right\|_2^2 .
\]

\Statex \Comment{Joint optimization}
\State Update the trainable parameters using
\[
    \mathcal{L}
    =
    \mathcal{L}_{\mathrm{video}}
    +
    \mathcal{L}_{\mathrm{action}}.
\]
\end{algorithmic}
\end{algorithm}

\begin{algorithm}[t]
\caption{Light-WAM inference}
\label{alg:lightwam_inference}
\begin{algorithmic}[1]
\Require Current observation $I_t$, language embedding $c$, proprioceptive state $p_t$
\Require Selected adapter layers $\mathcal{I}=\{8,16,24\}$
\State Build the current observation image from the camera inputs and encode it into a single-frame latent $z_{\mathrm{act}}$.
\State Build the cross-attention context
\[
    C=[c_1,\ldots,c_{128},c_{\mathrm{prop}}],
\]
where $c_{\mathrm{prop}}$ is obtained by projecting $p_t$.
\State Run one adapted video-backbone forward pass and collect selected adapted states:
\[
    \mathcal{H}
    =
    \{H_{\ell}\}_{\ell\in\mathcal{I}}
    =
    h_{\theta}(z_{\mathrm{act}},C).
\]
\State Predict the action chunk:
\[
    \hat{A}
    =
    \{\hat{a}_k\}_{k=0}^{K-1}
    =
    \pi_{\phi}(\mathcal{H}).
\]
\State Execute the predicted actions.
\end{algorithmic}
\end{algorithm}

\section{Training and Implementation Details}
\label{app:training_details}

\paragraph{Training setup.}
We train Light-WAM with AdamW, using a learning rate of $1\times10^{-4}$, weight decay of $1\times10^{-2}$, and a cosine learning-rate schedule with 1,000 warmup steps. 
All models are trained on 4 NVIDIA H100 GPUs. 
For LIBERO, we use a global batch size of 64. 
For RoboTwin 2.0, we use a global batch size of 128. 
Training uses cached Wan2.1 VAE latents to remove online VAE encoding from the training loop, while evaluation uses online VAE encoding.
The video backbone weights are frozen, and the trainable components include the backbone LoRA modules, WAM adapters, video prediction head, proprio encoder, and StateFusionActionExpert.

\paragraph{Checkpoint selection.}
For LIBERO, we select checkpoints for each suite: 60K steps for Spatial and Goal, 12.5K steps for Object, and 80K steps for Long. 
For RoboTwin 2.0, we evaluate the model trained for 460K steps.

\paragraph{Parameter breakdown.}
Table~\ref{tab:app_param_breakdown} reports the parameter composition of the default Light-WAM model. 
The model has 1.99B total parameters, of which 0.44B are trainable. 
Most trainable parameters come from the StateFusionActionExpert and backbone LoRA modules, while the pretrained video backbone and VAE remain frozen.

\clearpage

\begin{table}[t]
\centering
\caption{
Parameter breakdown of Light-WAM.
Numbers are reported in millions of parameters.
}
\label{tab:app_param_breakdown}
\setlength{\tabcolsep}{5pt}
\begin{tabular}{lccc}
\toprule
Component & Total & Trainable & Frozen \\
\midrule
Frozen video backbone & 1418.90M & 0.00M & 1418.90M \\
Backbone LoRA modules & 87.49M & 87.49M & 0.00M \\
WAM adapters & 2.37M & 2.37M & 0.00M \\
Video prediction head & 0.10M & 0.10M & 0.00M \\
StateFusionActionExpert & 351.03M & 351.03M & 0.00M \\
Proprio encoder & 0.04M & 0.04M & 0.00M \\
Wan VAE & 126.89M & 0.00M & 126.89M \\
\midrule
\textbf{Total} & \textbf{1986.82M} & \textbf{441.03M} & \textbf{1545.79M} \\
\bottomrule
\end{tabular}
\end{table}

\section{Full RoboTwin 2.0 Results}
\label{app:robotwin_full}


\begin{scriptsize}
\setlength{\tabcolsep}{3pt}
\begin{longtable}{lcccccccc}
\caption{
\textbf{Full RoboTwin 2.0 per-task results.}
}
\label{tab:app_robotwin_full}
\\
\toprule
\multirow{2}{*}{Task} 
& \multicolumn{2}{c}{$\pi_{0.5}$} 
& \multicolumn{2}{c}{X-VLA} 
& \multicolumn{2}{c}{Fast-WAM} 
& \multicolumn{2}{c}{Light-WAM} \\
\cmidrule(lr){2-3}
\cmidrule(lr){4-5}
\cmidrule(lr){6-7}
\cmidrule(lr){8-9}
& Clean & Rand. & Clean & Rand. & Clean & Rand. & Clean & Rand. \\
\midrule
\endfirsthead

\toprule
\multirow{2}{*}{Task} 
& \multicolumn{2}{c}{$\pi_{0.5}$} 
& \multicolumn{2}{c}{X-VLA} 
& \multicolumn{2}{c}{Fast-WAM} 
& \multicolumn{2}{c}{Light-WAM} \\
\cmidrule(lr){2-3}
\cmidrule(lr){4-5}
\cmidrule(lr){6-7}
\cmidrule(lr){8-9}
& Clean & Rand. & Clean & Rand. & Clean & Rand. & Clean & Rand. \\
\midrule
\endhead

\midrule
\multicolumn{9}{r}{Continued on next page} \\
\endfoot

\bottomrule
\endlastfoot

Adjust Bottle & 100 & 99 & 100 & 99 & 100 & 100 & 100 & 100 \\
Beat Block Hammer & 96 & 93 & 92 & 88 & 99 & 97 & 83 & 80 \\
Blocks Ranking RGB & 92 & 85 & 83 & 83 & 100 & 100 & 96 & 91 \\
Blocks Ranking Size & 49 & 26 & 67 & 74 & 94 & 98 & 57 & 54 \\
Click Alarmclock & 98 & 89 & 99 & 99 & 100 & 100 & 100 & 100 \\
Click Bell & 99 & 66 & 100 & 100 & 100 & 100 & 100 & 100 \\
Dump Bin Bigbin & 92 & 97 & 79 & 77 & 97 & 96 & 81 & 75 \\
Grab Roller & 100 & 100 & 100 & 100 & 100 & 100 & 100 & 98 \\
Handover Block & 66 & 57 & 73 & 37 & 95 & 81 & 71 & 59 \\
Handover Mic & 98 & 97 & 0 & 0 & 99 & 100 & 90 & 94 \\
Hanging Mug & 18 & 17 & 23 & 27 & 58 & 62 & 25 & 17 \\
Lift Pot & 96 & 85 & 99 & 100 & 100 & 100 & 93 & 93 \\
Move Can Pot & 51 & 55 & 89 & 86 & 90 & 88 & 57 & 74 \\
Move Pillbottle Pad & 84 & 61 & 73 & 71 & 100 & 99 & 69 & 74 \\
Move Playingcard Away & 96 & 84 & 93 & 98 & 100 & 100 & 93 & 92 \\
Move Stapler Pad & 56 & 42 & 78 & 73 & 77 & 64 & 26 & 34 \\
Open Laptop & 90 & 96 & 93 & 100 & 98 & 100 & 91 & 97 \\
Open Microwave & 34 & 77 & 79 & 71 & 62 & 45 & 76 & 59 \\
Pick Diverse Bottles & 81 & 71 & 58 & 36 & 80 & 85 & 61 & 57 \\
Pick Dual Bottles & 93 & 63 & 47 & 36 & 100 & 96 & 90 & 63 \\
Place A2B Left & 87 & 82 & 48 & 49 & 95 & 93 & 84 & 83 \\
Place A2B Right & 87 & 84 & 36 & 36 & 93 & 99 & 89 & 85 \\
Place Bread Basket & 77 & 64 & 81 & 71 & 91 & 93 & 81 & 80 \\
Place Bread Skillet & 85 & 66 & 77 & 67 & 90 & 93 & 92 & 82 \\
Place Burger Fries & 94 & 87 & 94 & 94 & 96 & 99 & 95 & 98 \\
Place Can Basket & 62 & 62 & 49 & 52 & 71 & 69 & 57 & 57 \\
Place Cans Plasticbox & 94 & 84 & 97 & 98 & 99 & 96 & 37 & 68 \\
Place Container Plate & 99 & 95 & 97 & 95 & 96 & 100 & 99 & 94 \\
Place Dual Shoes & 75 & 75 & 79 & 88 & 94 & 88 & 53 & 51 \\
Place Empty Cup & 100 & 99 & 100 & 98 & 100 & 100 & 91 & 95 \\
Place Fan & 87 & 85 & 80 & 75 & 96 & 96 & 77 & 74 \\
Place Mouse Pad & 60 & 39 & 70 & 70 & 83 & 89 & 58 & 62 \\
Place Object Basket & 80 & 76 & 44 & 39 & 89 & 88 & 81 & 72 \\
Place Object Scale & 86 & 80 & 52 & 74 & 90 & 97 & 72 & 74 \\
Place Object Stand & 91 & 85 & 86 & 88 & 90 & 94 & 78 & 85 \\
Place Phone Stand & 81 & 81 & 88 & 87 & 97 & 99 & 85 & 88 \\
Place Shoe & 92 & 93 & 96 & 95 & 96 & 99 & 84 & 87 \\
Press Stapler & 87 & 83 & 92 & 98 & 90 & 97 & 65 & 76 \\
Put Bottles Dustbin & 84 & 79 & 74 & 77 & 95 & 90 & 65 & 65 \\
Put Object Cabinet & 80 & 79 & 46 & 48 & 94 & 89 & 80 & 68 \\
Rotate QRcode & 89 & 87 & 34 & 33 & 93 & 89 & 72 & 84 \\
Scan Object & 72 & 65 & 14 & 36 & 89 & 92 & 60 & 52 \\
Shake Bottle Horizontally & 99 & 99 & 100 & 100 & 100 & 100 & 100 & 98 \\
Shake Bottle & 99 & 97 & 99 & 100 & 100 & 100 & 100 & 99 \\
Stack Blocks Three & 91 & 76 & 6 & 10 & 95 & 97 & 65 & 67 \\
Stack Blocks Two & 97 & 100 & 92 & 87 & 100 & 100 & 94 & 91 \\
Stack Bowls Three & 77 & 71 & 76 & 86 & 80 & 81 & 65 & 72 \\
Stack Bowls Two & 95 & 96 & 96 & 93 & 92 & 98 & 91 & 97 \\
Stamp Seal & 79 & 55 & 76 & 82 & 90 & 94 & 60 & 63 \\
Turn Switch & 62 & 54 & 40 & 61 & 61 & 59 & 33 & 39 \\
\midrule
\textbf{Average} & 82.7 & 76.8 & 72.9 & 72.8 & 91.9 & 91.8 & 76.4 & 76.3 \\

\end{longtable}
\end{scriptsize}

\clearpage

\section{Real-World Rollouts}
\label{app:real_world_rollouts}

\begin{figure}[H]
    \centering
    \includegraphics[width=0.85\textwidth]{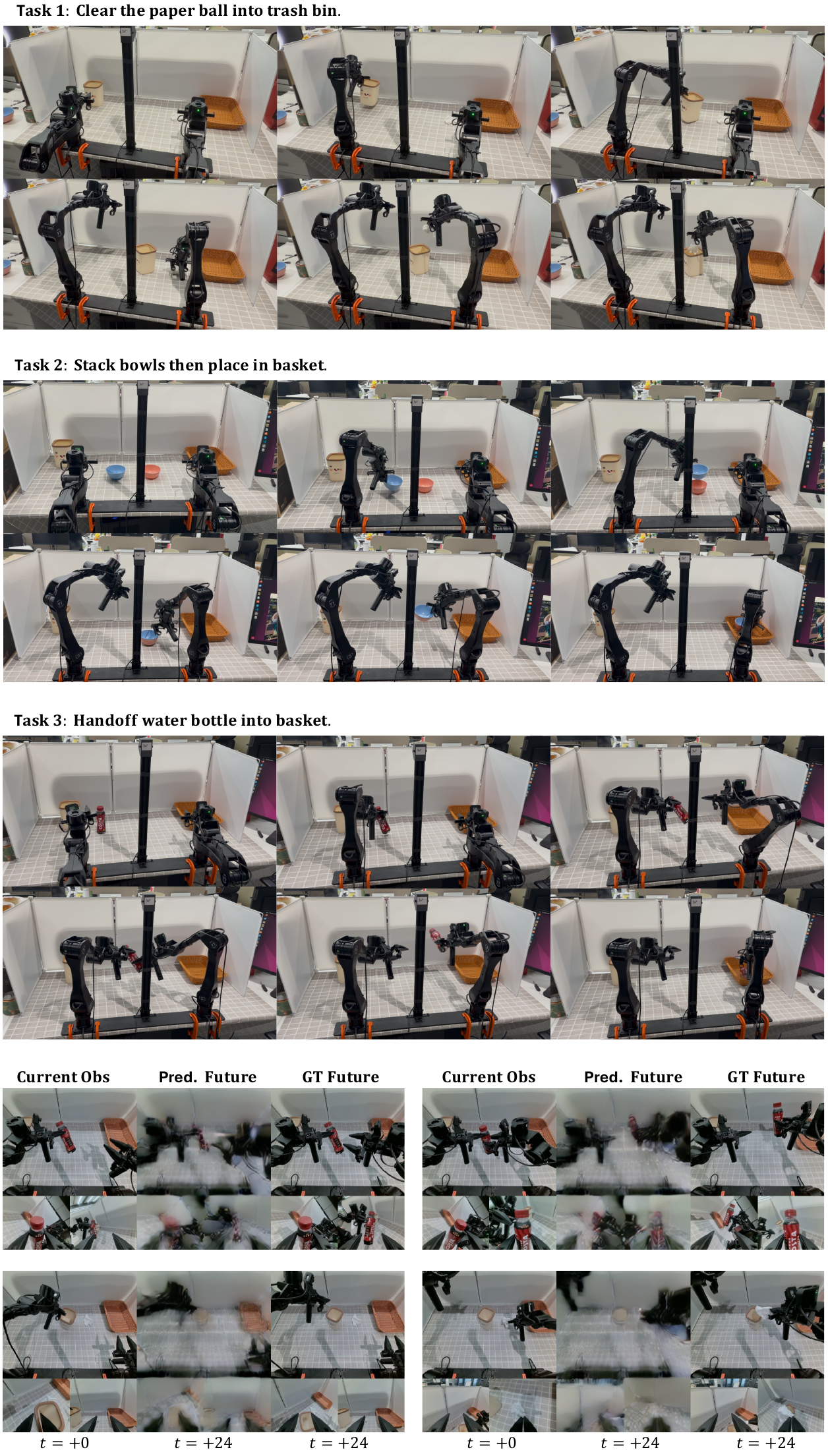}
    \caption{
    \textbf{Additional real-world rollouts.}
    Rollout frames on three dual-arm tasks and future-video predictions compared with ground-truth future frames.
    }
    \label{fig:real_world_rollouts}
\end{figure}

\end{document}